\documentclass[runningheads]{llncs}
\usepackage[T1]{fontenc}
\usepackage{cite}
\usepackage{amsmath,amssymb,amsfonts}
\usepackage{algorithmic}
\usepackage{graphicx}
\usepackage{textcomp}
\usepackage{multirow}
\usepackage{comment}
\usepackage{url}
\usepackage{wrapfig}
\usepackage{subcaption}
\usepackage{diagbox}
\usepackage{url}
\usepackage{arydshln}
\usepackage{nicematrix}
\usepackage{tikz}
\usepackage{calc}
\usepackage{booktabs}
\usepackage{xcolor,colortbl}
\usepackage{graphicx}
\usepackage{makecell}
\usepackage{nicematrix}
\usepackage{hyperref}
\usepackage[flushleft]{threeparttable}

\begin{document}

\title{EncodeNet: A Framework for Boosting DNN Accuracy with Entropy-driven Generalized Converting Autoencoder}

\author{
Hasanul Mahmud
\and
Kevin Desai
\and
Palden Lama
\and
Sushil K. Prasad
}

\authorrunning{H. Mahmud et al.}
\titlerunning{EncodeNet}
%
\institute{
The University of Texas at San Antonio\\
\email{\{hasanul.mahmud, kevin.desai, palden.lama, sushil.prasad\}@utsa.edu}
}
\maketitle
\begin{abstract}

Image classification is a fundamental task in computer vision, and the quest to enhance DNN accuracy without inflating model size or latency remains a pressing concern. We make a couple of advances in this regard, leading to a novel EncodeNet design and training framework. The first advancement involves Converting Autoencoders, a novel approach that transforms images into an easy-to-classify image of its class. Our prior work that applied the Converting Autoencoder and a simple classifier in tandem achieved moderate accuracy over simple datasets, such as MNIST and FMNIST. However, on more complex datasets like CIFAR-10, the Converting Autoencoder has a large reconstruction loss, making it unsuitable for enhancing DNN accuracy. To address these limitations, we generalize the design of Converting Autoencoders by leveraging a larger class of DNNs, those with architectures comprising feature extraction layers followed by classification layers. We incorporate a generalized algorithmic design of the Converting Autoencoder and intraclass clustering to identify representative images, leading to optimized image feature learning. Next, we demonstrate the effectiveness of our EncodeNet design and training framework, improving the accuracy of well-trained baseline DNNs while maintaining the overall model size. EncodeNet's building blocks comprise the trained encoder from our generalized Converting Autoencoders transferring knowledge to a lightweight classifier network - also extracted from the baseline DNN. Our experimental results demonstrate that EncodeNet improves the accuracy of VGG16 from 92.64\% to 94.05\% on CIFAR-10, and RestNet20 from 74.56\% to 76.04\% on CIFAR-100. It outperforms state-of-the-art techniques that rely on knowledge distillation and attention mechanisms, delivering higher accuracy for models of comparable size.

\end{abstract}

\section{Introduction}
\label{Introduction}

In recent years, deep learning models have gained tremendous success in many computer vision tasks such as image classification~\cite{ResNet,Hu2017SqueezeandExcitationN}, object detection~\cite{He2017MaskR, Ren2015FasterRT}, and semantic segmentation~\cite{Zhao2016PyramidSP,Shelhamer2017FullyCN}. However, this also brings the challenge of creating models that strike a balance between accuracy and efficiency. Although larger DNN models exhibit high accuracy in computer vision tasks, they are computationally expensive. Therefore, there is a growing interest in designing efficient DNNs to achieve high accuracy while maintaining low computational cost. One of the approaches to achieve this goal is model compression, which involves pruning and quantization of large models while preserving their accuracy to some extent~\cite{pruningsurvey2021,thinnetpruning,Knowledgewithpruning,Energywareprune,sparse_representation,Jacob2017QuantizationAT, Louizos2017BayesianCF}. 
Alternatively, techniques like knowledge distillation \cite{Knowledgedistillation_hinton, Beyer-CVPR22, Factor_Transfer, romero2015fitnets}, and attention mechanism \cite{park2018bam,woo2018cbam,Liu2021GlobalAM} focus on improving the accuracy of baseline models with either minimal or no increase in the model size. Our work is in line with the latter approach. 

In our prior work~\cite{mahmud2024converting}, we introduced a Converting Autoencoder (CAE) that transforms input images into easy-to-classify images of its class and subsequently processed the transformed images with a simple classifier. The CAE model, which was based on the U-Net architecture relied on early-exiting DNN framework for its training and achieved moderate accuracy over simple datasets such as MNIST, FMNIST, etc. However, its accuracy on more complex datasets like CIFAR-10 was limited to 78-80\% when passed to AlexNet or ResNet50 for inference. In this paper, we generalize the design of our Converting Autoencoder by systematically deriving its structure from given baseline DNN, making it applicable to a large class of DNNs and more complex datasets. Leveraging the Generalized Converting Autoencoder (GCAE), we developed \emph{EncodeNet}, a novel integrative framework that enhances the accuracy of any baseline DNN with a modular architecture of feature extraction layers followed by classification layers, achieving performance on par with significantly larger models. Our framework surpasses competing techniques, including state-of-the-art knowledge distillation and attention mechanism-based methods.


\emph{EncodeNet} involves two stages of model training. In the initial stage, a converting autoencoder is trained to transform an image into a representative image within the same class, thus extracting its salient features. To identify the most representative images, each class of images is grouped into different clusters based on their similarity. For each cluster, the image that can classified with the lowest entropy using the baseline DNN is selected as the representative image. A low entropy of a classification result indicates a high confidence in the prediction. Intraclass clustering effectively reduces the reconstruction loss of the converting autoencoder. 
In the second stage, we combine the encoder layers of the trained converting autoencoder with additional layers and filters derived from the classification layers of the baseline DNN model. We train the new DNN by freezing the pre-trained encoder layers and only training the remaining layers of the network. By doing so, we leverage the learned representations from the autoencoder and fine-tune them for image classification. To our knowledge, this is the first integrative framework designed for entropy-driven representative feature extraction with the help of a Generalized Converting Autoencoder. It synergizes its capabilities with a thin subnetwork extracted from a baseline DNN, yielding an equivalent DNN with significantly improved accuracy.

Our key contributions are as follows:
\begin{itemize}
    \item We developed an algorithmic approach to generalize the design of our Converting Autoencoder from U-Net architecture and early-exiting DNNs to a larger class of DNNs and more complex datasets than previously possible. 
       
    \item We designed \emph{EncodeNet}, a new framework leveraging Generalized Converting Autoencoder for training lightweight DNNs that can achieve accuracy comparable to significantly larger models without increasing model size. Our approach competes well with state-of-the-art techniques, including  Knowledge Distillation and Attention Mechanism, due to its versatility. It can be implemented even when large teacher models are unavailable, yet it still attains comparable or greater accuracy.
     
    \item 
    Experimental results using CIFAR-10 and CIFAR-100 datasets demonstrate the remarkable effectiveness of our approach, outperforming competing techniques. EncodeNet improves the accuracy of VGG16 from 92.64\% to 94.05\% on CIFAR-10, and RestNet20 from 74.56\% to 76.04\% on CIFAR-100. It outperforms KD (Knowledge Distillation)~\cite{Knowledgedistillation_hinton}, RKD (Relational Knowledge Distillation)~\cite{RKD}, FitNet~\cite{romero2015fitnets}, and FT (Factor Transfer)~\cite{Factor_Transfer} for both ResNet and VGG networks. EncodeNet enhances the accuracy of ResNet50 on CIFAR-100 from 77.23\% to 80.1\%, outperforming attention mechanims based techniques, Squeeze-and-Excitation Networks (SE) ~\cite{hu2018squeeze} and Bottleneck Attention Module (BAM)~\cite{park2018bam}. It achieves comparable accuracy with  Convolutional block attention module (CBAM) \cite{woo2018cbam}, and Global attention mechanism (GAM) \cite{Liu2021GlobalAM} while maintaining relatively small model size. 

\end{itemize}

\section{Related Work}
\label{related_work}
In this section, we review some of the significant methods that focus on improving DNN accuracy with either minimal or no increase in the model size.

\subsection{Autoencoders}
An autoencoder is an artificial neural network that learns efficient encodings of unlabeled data. It consists of an Encoder, which learns how to encode data into a reduced representation efficiently, and a Decoder, which learns how to reconstruct the data back to a representation that is as close to the original input as possible~\cite{vincent2010stacked}. There are several variations of autoencoders. The denoising autoencoders are trained to recover original input from intentionally perturbed or noisy input\cite{Goodfellow-et-al-2016}, with the aim to learn a more robust representation of input data. 
A variational autoencoder is a generative model that can produce different variations of existing data samples~\cite{vae}. Converting autoencoders\cite{mahmud2024converting} has been used to improve the performance of the early-exiting framework. However, this approach is limited to simple networks and datasets. In addition, it relies on early-exiting frameworks, further limiting its capability. Autoencoders are also widely used for dimensionality reduction, denoising, data augmentation, and anomaly detection. What sets us apart is our synergistic system approach integrating an autoencoder trained for representative image transformation, transferring its knowledge to a thin subnetwork of a baseline network yielding lightweight DNNs for resource-constrained devices.

\subsection{Knowledge Distillation}

Initially proposed by Hinton et al.~\cite{Knowledgedistillation_hinton}, knowledge distillation is a model compression method that distills the knowledge in a large network or ensemble of networks into a small student network by forcing the student’s predictions to match those of the teacher. The effectiveness of distillation has been demonstrated in many studies, e.g. \cite{Cho-ICCV19,mishra2017apprentice,Romero2014}, using various student and teacher architecture patterns with different depths and widths. 
More recently, Beyer et al.~\cite{Beyer-CVPR22} identified that aggressive data augmentation and a long training schedule can drastically improve the effectiveness of the original knowledge distillation method proposed in~\cite{Knowledgedistillation_hinton}. Our framework presents an alternative method that results in a lightweight model, ideal for devices with limited resources. It does not depend on large models, yet it surpasses state-of-the-art methods based on knowledge distillation.

\subsection{Streamlined DNN Architectures}

Several works have proposed methods to streamline DNN architectures for resource-constrained platforms. Adadeep~\cite{Adadeep} is a usage-driven, automated DNN compression framework that systematically explores the trade-off between performance and resource constraints. SubFlow~\cite{subflow} uses subnetwork pruning to find the optimal subnetworks for each layer that can preserve the accuracy of the original network, and fine-tunes them with a sparsity regularization term. MobileNet~\cite{mobilenets} and MobileNetV2~\cite{mobilenetv2} are DNN architectures designed for mobile and embedded vision applications, which, respectively, use depth-wise separable convolutions and inverted residual blocks with bottle-necking features to reduce the number of parameters and computations. MobileNetV2 also introduces linear bottlenecks and shortcut connections to improve the model's efficiency and accuracy. Unlike these approaches, our work focuses on enhancing the accuracy of existing DNN models without introducing new network architectures. 

\subsection{Attention Mechanisms}

Several studies have focused on performance improvements using attention mechanisms for image classification tasks. 
The attention mechanism is a computational technique enabling neural networks to concentrate on pertinent input data segments during task execution. This mirrors the selective attention mechanism observed in human cognition, allowing models to assign varying levels of significance to distinct segments of the input sequence. Squeeze-and-Excitation Networks (SENet)\cite{hu2018squeeze} is the first to use channel attention and channel-wise-feature-fusion to suppress the unimportant channels. The convolutional block attention module (CBAM)\cite{woo2018cbam} places the channel and spatial attention operation sequentially, while the bottleneck attention module (BAM)\cite{park2018bam} did it in parallel. Global attention\cite{Liu2021GlobalAM} mechanism was proposed that boosts the performance of deep neural networks by reducing information reduction. Even though attention mechanisms have been successful in increasing DNN performance, additional parameters are needed to learn how to weigh the importance of different parts of the input sequences. 
\section{Method} \label{Implementation}

\begin{figure*}[htb]
      \centering
    \includegraphics[width=0.95\textwidth]{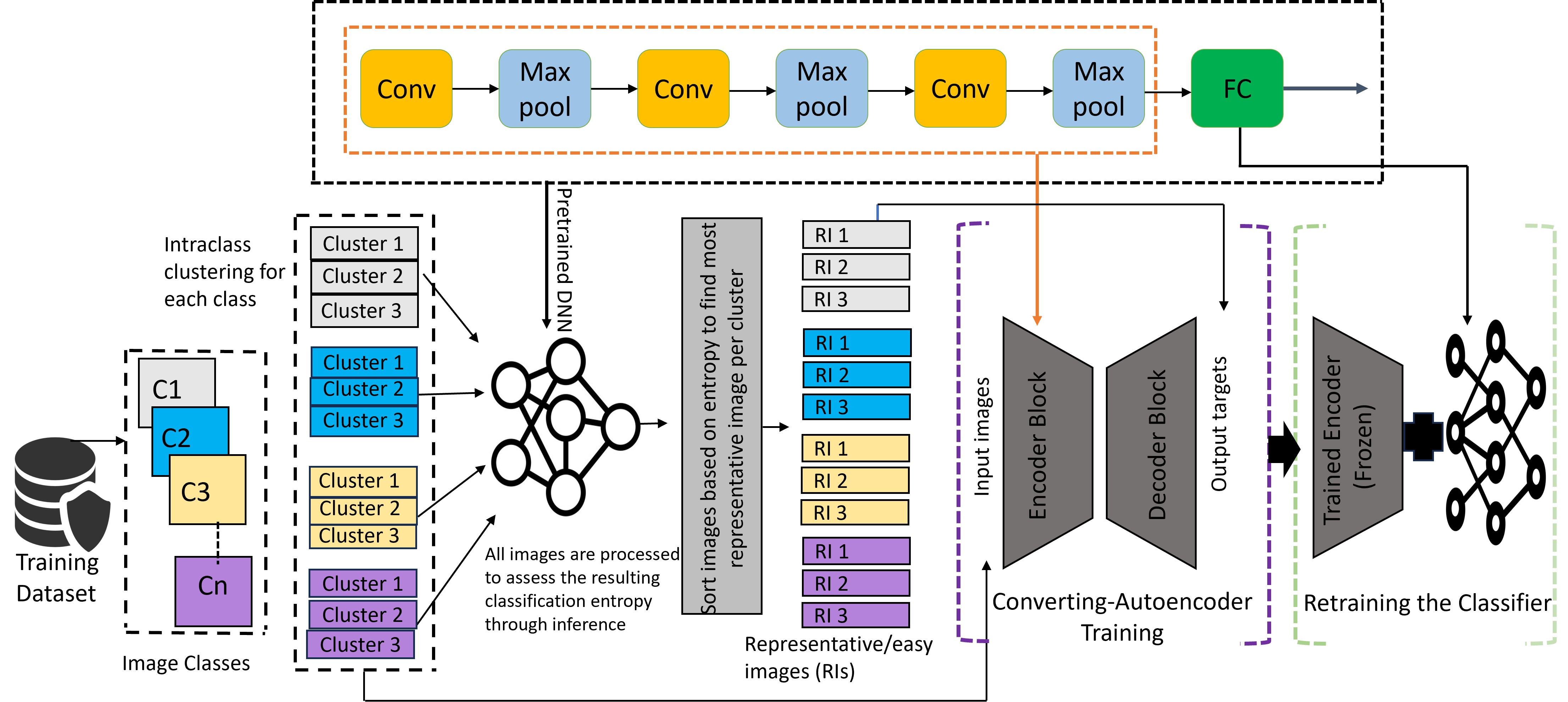}
  \caption{Overview of the EncodeNet Framework: Intraclass clustering of images, followed by sorting them according to their classification entropy to pinpoint the most representative image for each cluster. Subsequently, a converting autoencoder is trained to transform input images into their corresponding representative images within the same class and cluster. Finally, the trained encoder layers are detached from the autoencoder, coupled with fully connected layers, and retrained to perform classification.}
  \label{fig:framework}
\end{figure*}

In this section, we introduce the details of our proposed EncodeNet framework, which comprises of three key aspects - (1) generalizing the Converting Autoencoders for efficient training setup, (2) intraclass clustering and entropy-based image selection to support representative feature learning with Converting Autoencoder, and (3) knowledge transfer from Converting Autoencoder for image classification.  
Figure~\ref{fig:framework} provides the details of the framework, with more details provided in the sections below.

\subsection{Generalized Converting Autoencoder}
\subsubsection{Converting Autoencoder Design:} 
 Autoencoders are widely employed for various tasks such as image reconstruction, image compression, denoising, dimensionality reduction, etc. In our attempt, we crafted an autoencoder to transform images into easily classifiable representations within their respective classes. Designing a lightweight autoencoder poses challenges, particularly when the encoder component can not successfully extract features, potentially leading to miss-transformations. While the UNet~\cite{Ronneberger2015UNetCN} model offers a built-in architecture for encoder and decoder blocks, it fails to convert images into easy representations. Thus, our primary objective was developing an encoder capable of accurately capturing features and designing the decoder block for efficient transformation. In general, DNN architecture for the image classification task consisted mainly of feature extraction and classification layers. The DNN's feature extraction stage plays a crucial role in capturing vital features and reducing the size of feature maps. The classification layers utilize these identified features to make predictions. Considering this, our framework commences by selecting a baseline for autoencoder design using the feature extraction layers as the encoder of the autoencoder, as depicted in Figure ~\ref{fig:framework}. Specifically, our focus is on the feature extractor component of the Deep Neural Network (DNN), with the exclusion of classification layers. Next, we create the corresponding decoder block, a meticulous process aimed at complementing the feature extraction mechanism and efficiently reconstructing the input data. Various architectures were employed to optimize the decoder block, aiming to reduce the autoencoder's reconstruction loss. Convolutional layers and Upsampling layers were utilized to upscale the extracted feature maps to their original dimensions. By discarding classification layers in this phase, our objective is to design a customized autoencoder capable of capturing and representing crucial features from the input data, thereby paving the way for subsequent improvements in model performance.

\subsection{Representative Feature Learning with Converting Autoencoder}\label{intraclass-clustering}
\subsubsection{Intraclass Clustering: }
\begin{figure}[htb]
\centering
    \includegraphics[width=0.5\textwidth]{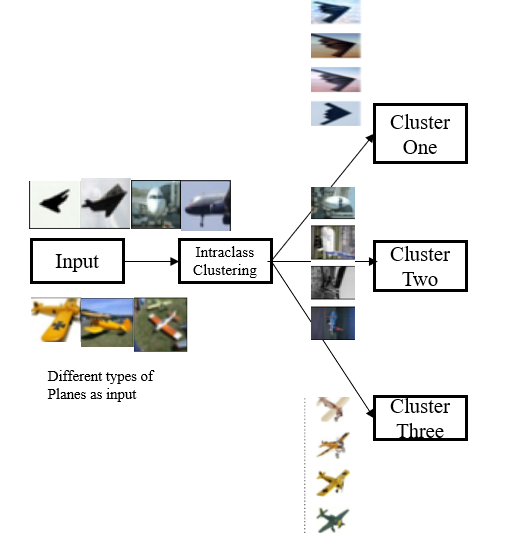}
    \caption{Intraclass clustering on airplane class of CIFAR-10.}
    \label{fig:Clustering}
\end{figure} 
Complex datasets such as CIFAR-10, CIFAR-100 can have high dissimilarity among images of the same class, which poses a challenge for representative image transformation with Converting Autoencoders. For example, Fig.~\ref{fig:Clustering} shows that the airplane class in CIFAR-10 dataset contains various types of airplanes (e.g., fighter jet, commercial airplane, etc.) that differ significantly from each other. It is more difficult and costly to transform a fighter jet into a representative image of a commercial airplane than into a representative image of a fighter jet. To tackle this problem, we introduce intraclass clustering that groups images of the same class into different clusters based on their similarity. This facilitates effective training for our Converting Autoencoder to perform the representative image transformation within each cluster. Table ~\ref{tab:Reconstruction_loss} shows the effectiveness of intraclass clustering in minimizing the reconstruction loss for the conversion of the images.

\begin{tt}
\begin{table}[htb]
    \centering
    \caption{Impact of intraclass clustering in minimizing the reconstruction loss on CIFAR-10 datasets after 500 epochs of training}
    \begin{tabular}{c@{\hskip 5mm}c@{\hskip 5mm}c}
    \hline
    \multicolumn{3}{c } {Reconstruction loss of converting autoencoder} \\
        \hline
        
                   Models & Without Clustering & With Clustering \\ 
        \hline
             VGG8 & 0.019 & 0.008 \\
             VGG16 & 0.013 & 0.07  
            \\    

            ResNet18 & 0.016 & 0.009   
            \\ 
            ResNet20 & 0.025 & 0.0107   
            \\ 
    \hline
    \end{tabular}
    \label{tab:Reconstruction_loss}
\end{table}
\end{tt}

For efficient clustering, we extracted important features of the input image using a VGG-16\cite{VGG16} model trained on ImageNet\cite{imagenet} data. 
The pre-trained VGG16 model is a convolutional neural network (CNN) model that has already been trained on a large dataset, typically for image recognition tasks. 
In this case, the input image to the VGG16 model had a shape of (32,32,3), which means it had a width and height of 32 pixels and 3 color channels (red, green, and blue). When using VGG16 for feature extraction, it's common to remove the last few layers of the network, including the fully connected and classification layers, and use the remaining layers as a feature extractor. We fed input images to VGG16 and used the 13th layer to extract a feature map with a shape of (1,1,512). This reduced shape makes it more suitable for applying a clustering algorithm, which aims to group similar data points together based on their features. We applied three different clustering techniques to group similar images together within each class: k-means clustering, HDBSCAN, and cosine similarity clustering. Among these three techniques, k-means clustering provided the best results. 

\begin{figure}[htb]
    \centering
    \includegraphics[width=0.6\textwidth]{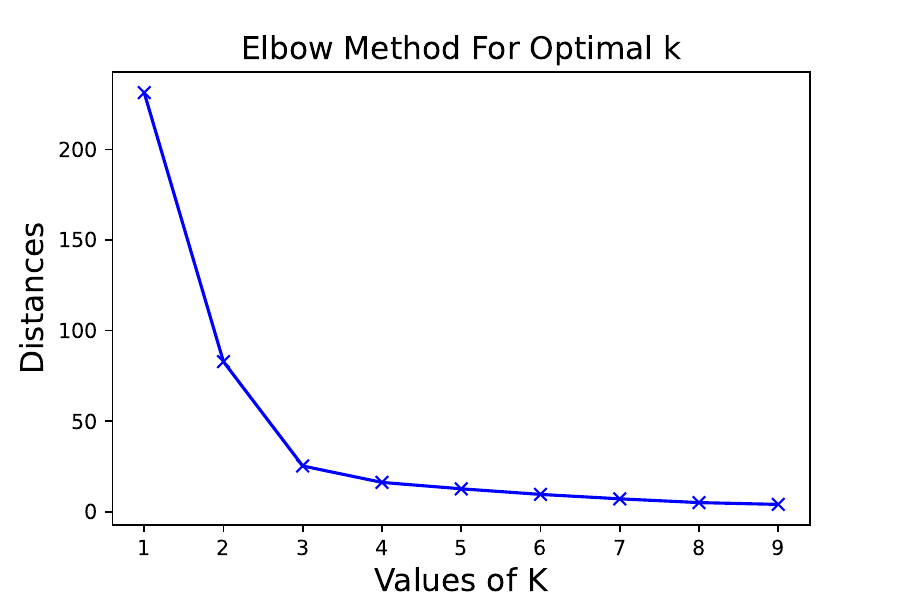}

    \caption{Finding the optimal number of clusters.}
    \label{fig:Clusters_Optimal}
\end{figure}

\begin{figure}[htb]
    \centering
    \includegraphics[width=0.7\textwidth]{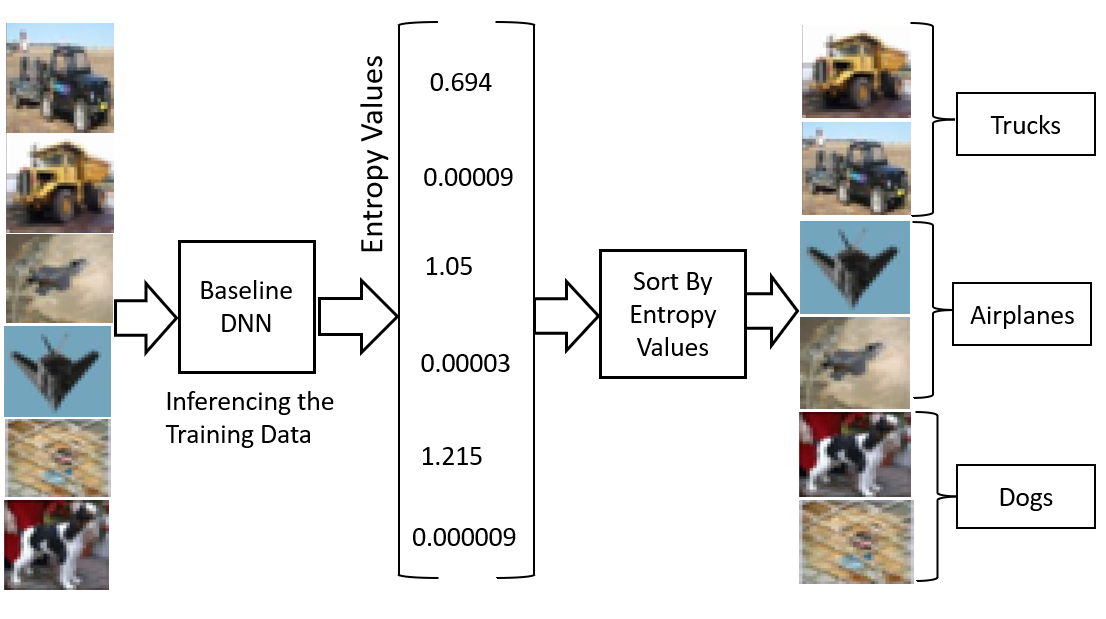}
    \caption{Measuring the entropy of baseline DNN model.}
    \label{fig:labeling}
\end{figure}
To determine the optimal number of clusters for k-means clustering, we used the Elbow method~\cite{Elbow}, which is a technique that plots the sum of squared distances between data points and their assigned cluster centers for different values of k (the number of clusters). The “elbow” point on this plot represents the value of k where the decrease in sum of squared distances begins to level off. As shown in Figure~\ref{fig:Clusters_Optimal}, we found that the elbow point was at k=3. We aimed to group similar images together into 3 distinct clusters within each class of the CIFAR10 dataset.

\subsubsection{Entropy-based Representative Image Extraction: }

The level of difficulty in the image-classification task and the requirement for complex DNN models depend on the complexity of the input. We measure the confidence of a classification result produced by a baseline DNN model (e.g., VGG,ResNet) based on the entropy of the prediction. If an input image is predicted to belong to a single class with 100\% probability, the entropy is zero, as there is no uncertainty in the classification. If the prediction probability is equally distributed across all classes, the entropy is at its maximum. As shown in Fig.~\ref{fig:labeling}, we first sort the images in each cluster and class by their entropy. The representative image is identified as the one that has the lowest entropy in the corresponding class and cluster.
The next stage of the framework trains a Converting Autoencoder to transform any input image of a particular class and cluster into the corresponding representative image.

\subsubsection{Representative Feature Learning through Image Transformation: }

\begin{figure}[htb]
    \centering
    \begin{subfigure}[b]{0.7\textwidth}
    \centering
    \includegraphics[width=\textwidth]{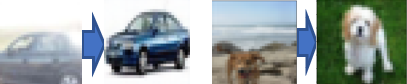}
    \end{subfigure}
    \caption{ A converting autoencoder transforming CIFAR10 images into the representative images of their corresponding class and cluster.}
    \label{fig:HardToEasy-CIFAR-10}
\end{figure}

We train our Converting Autoencoder to encode any given input image into an efficient representation that can be decoded into the representative image belonging to the same class and cluster. 
We evaluated the effectiveness of our approach using the CIFAR-10 and CIFAR-100 datasets with and without intraclass clustering. The Converting Autoencoder's performance was evaluated in terms of reconstruction loss, which measures the similarity between the output and the target image. The target image is the easiest image within the cluster for a given input image belonging to a cluster. As a result of the intraclass clustering, the reconstruction loss of our autoencoder dropped by 46\%, from 0.0012 to 0.00065. Fig.~\ref{fig:HardToEasy-CIFAR-10} illustrates the transformation of the given images to the corresponding representative images using the autoencoder trained with CIFAR-10 dataset.

\subsection{Knowledge Transfer from Converting Autoencoder for Image Classification}

In the EncodeNet framework, the encoder layers of an autoencoder captures the essential features of the given input image and the representative image that can aid in efficient image classification. Therefore, we employ transfer learning to train a DNN model that consists of the pre-trained encoder layers and additional layers obtained through the classification part of a baseline DNN model. By freezing the pre-trained encoder layers and only training the remaining layers of the network, we utilize the learned representations from the Converting Autoencoder and adapt them for the task of image classification. This approach integrates the Converting Autoencoder's representative image transformation capability into an efficient DNN model with enhanced accuracy.

\section{Experiments \& Results}
In this section, we discuss the experiments we conducted on the different datasets to evaluate our proposed EncodeNet framework against other competing techniques.
We first provide the implementation details of our work and present the datasets we have used in our experiments.
Next, we provide results demonstrating the enhanced performance achieved for the baseline DNNs as a result of our EncodeNet framework.
We also conduct ablation studies to evaluate the importance of our proposed use of intraclass clustering and the auto-encoder framework.
Lastly, we compare our approach against widely used methods for Knowledge Distillation and Attention mechanisms that aim at enhancing the performance of the given DNN model.

\subsection{Implementation details}
Our model is built on the TensorFlow framework~\cite{Abadi2016TensorFlowAS} and trained on two datasets: CIFAR-10 and CIFAR-100 as described in  ~\ref{datasets}. For the sake of comparison with Knowledge Distillation frameworks, we utilized the repository \cite{mdistiller} that was built on the PyTorch framework~\cite{Paszke2019PyTorchAI}. For training the baseline DNN models, we set the learning rate $1\mathrm{e}{-1}$ and a weight decay of $1\mathrm{e}{-4}$. We conducted training sessions for the baseline models over 300 epochs. For the autoencoder aimed at converting images into simpler representations, an extended training period of 500 epochs was necessary to minimize loss effectively. The optimization objective for the Converting Autoencoders centered on minimizing the reconstruction loss measured in terms of mean squared error (MSE). 

\subsection{Datasets} \label{datasets}
We evaluated our proposed framework on two widely used image classification datasets: CIFAR-10~\cite{cifar10}, CIFAR-100~\cite{Cifar100}. The CIFAR-10 dataset contains 60,000 color images of size 32x32, of which 50,000 are used for training and 10,000 for testing. The dataset includes 10 different classes of images, consisting of airplanes, cars, birds, cats, deer, dogs, frogs, horses, ships, and trucks. Whereas the CIFAR-100 dataset has 100 classes containing 600 images each, of which 500 are used for training and 100 for testing.

\subsection{Improvement of Baseline DNN}

\begin{tt}
\begin{table}[htb]
    \centering
    \caption{Results of our EncodeNet framework used to improve the accuracy (in \%) of different baseline DNN models on the CIFAR-10 and CIFAR-100 datasets.}
    \vspace{2mm}
    \begin{tabular}{l@{\hskip 5mm}c@{\hskip 3mm}c@{\hskip 5mm}c@{\hskip 3mm}c}
    \hline
        Architectures & \multicolumn{2}{c } {CIFAR-10} & \multicolumn{2}{c } {CIFAR-100} \\
        \cline{2-5} 
        & Baseline & \textbf{EncodeNet} & Baseline & \textbf{EncodeNet}\\ 
        \hline
            VGG8 & 89.25 & \textbf{91.60} & 70.23 & \textbf{73.41}   \\
            VGG16 & 92.64  & \textbf{94.06} & 73.37 & \textbf{75.11} \\
            ResNet18 & 91.12 & \textbf{92.87} & 74.22  & \textbf{75.29} \\
            ResNet20 & 92.56 & \textbf{93.64}  & 74.56 & \textbf{76.04}\\ 
    \hline
    \end{tabular}
    \label{tab:Improved_baselines}
\end{table}
\end{tt}

Table~\ref{tab:Improved_baselines} presents results illustrating the improvement in accuracy for the baseline network achieved through the robust training of our framework. For comparison, we evaluated various derivatives of two types of networks: VGG~\cite{VGG16} and ResNet~\cite{ResNet}. For example, VGG8 exhibited an increase in accuracy of 1.89\% and 3.2\% for the CIFAR-10 and CIFAR-100 datasets, respectively. Whereas, VGG16 demonstrated an improvement of approximately 1.8\% to 2.0\% in accuracy over the baseline due to EncodeNet.

\subsection{Ablation Studies}
\begin{table}[htb]
  \caption{Ablation studies for the EncodeNet framework on the CIFAR-10 dataset using ResNet18 as the baseline.}
  \label{tab:Ablation Studies}
  \vspace{2mm}
  \centering
  \begin{tabular}{l@{\hskip 5mm}c}
    \toprule
    Architecture Setup       & Accuracy ($\%$)  \\
    \midrule
    Baseline - ResNet18 without EncodeNet  & 91.12 \\
    ResNet 18 + no IC + AE-target is the same as input & 89.26 ($\downarrow$ 1.86\%)\\
    ResNet 18 + no IC + AE-lowest entropy image per class  & 92.11 ($\uparrow$ 0.99\%) \\
    ResNet 18 + IC + AE-representative image for each sub-class &  92.87 ($\uparrow$ 1.75\%) \\
    \hline
  \end{tabular}
\end{table}

To assess the impact of the different parts of our EncodeNet framework, we conducted ablation studies on the CIFAR-10 dataset using ResNet18 as the baseline DNN. Table ~\ref{tab:Ablation Studies} shows the said results. 
The baseline classification accuracy for ResNet18, without using EncodeNet, on CIFAR-10 is 91.12\%.
EncodeNet relies on an autoencoder for the approach to work. However, we can have different target image for feature learning.
In the first setup, we set the target image for the AE as the same as the input image. By using this setup, the classification accuracy reduces to 89.26\% (decrease of 1.86\%), indicating ineffective feature learning for the entire dataset.
In the second setup we perform entropy-based sorting of the image of a given class and use the lowest entropy image as the representative image for the entire class.
Doing so, EncodeNet provides improved learning of the image features, thereby, increasing the accuracy to 92.11\% (increase of 0.99\%).
We improve the feature learning and increase the accuracy even further with the introduction of intraclass clustering. Here, instead of using one representative image for the entire class, we derive different representative images for each of the sub-classes.
Specifically, the last setup, which is the complete EncodeNet framework, leads to the classification accuracy of 92.87\% (increase of 1.75\%).
The results for the ablation study validates the importance of the proposed autoencoder-based framework as well as the use of intraclass clustering.

\subsection{Comparison with Knowledge Distillation Techniques}

\begin{tt}
\begin{table}[htb]
    \centering
    \caption{Comparison with various Knowledge Distillation techniques on the CIFAR-100 dataset.}
    \vspace{2mm}
    \begin{tabular}{lc@{\hskip 5mm}c}
    \hline
                \centering Frameworks \hspace{2 mm}  
    \thead{Teacher:- \\ Student:-}   & \thead{ ResNet110 (74.56) \\   ResNet 32 (71.09)}   & \thead{VGG16 (73.35) \\  VGG8 (70.21)}       \\
        \hline
            Baseline(student) & 71.09 & 70.21   \\
            KD~\cite{Knowledgedistillation_hinton} & 72.94 & 72.52   \\
            RKD~\cite{RKD} & 71.03 &  71.48   \\
            FitNet~\cite{romero2015fitnets} & 71.39 & 71.08   \\
            FT~\cite{Factor_Transfer} & 73.47  & 71.76  \\
            \textbf{EncodeNet} & \textbf{74.13}  & \textbf{73.41} \\
    \hline
    \end{tabular}
    \label{tab:KD_Comparison}
\end{table}
\end{tt}

Knowledge Distillation approaches are known to increase the accuracy of a given baseline DNN model (Student) using a larger DNN model (Teacher).
This relies on the availability of a larger Teacher model, which may not always be the case.
Nonetheless, we compare our EncodeNet framework against four different knowledge techniques: KD (Knowledge Distillation)~\cite{Knowledgedistillation_hinton}, RKD (Relational Knowledge Distillation)~\cite{RKD}, FitNet~\cite{romero2015fitnets}, and FT (Factor Transfer)~\cite{Factor_Transfer}. 
Table \ref{tab:KD_Comparison} shows the results for these methods for two diferent student-teacher combinations on the CIFAR-100 dataset.
We employed two sets of teacher-student models: one with ResNet-110 as the teacher and ResNet-32 as the student, and the other with VGG16 as the teacher and VGG8 as the student. The results demonstrate that our method outperforms other knowledge distillation techniques for both ResNet and VGG networks.

\subsection{Comparison with Attention Mechanism based Techniques}

\begin{table}[ht!]
  \caption{Comparison with Attention Mechanism based techniques on the CIFAR-100 dataset.}
  \label{table:attention_mechanism}
  \centering
  \begin{tabular}{l@{\hskip 5mm}c@{\hskip 5mm}c}
    \toprule
    Architecture    & Parameters    & Accuracy ($\%$) \\ %
    \midrule
    ResNet 50~\cite{ResNet} & 23.71M  &  77.23   \\ %
    ResNet 50 + SE~\cite{hu2018squeeze} & 26.22M   & 79.71    \\ %
    ResNet 50 + BAM~\cite{park2018bam}  & 24.06M  & 80.03    \\ %
    ResNet 50 + CBAM~\cite{woo2018cbam} & 26.24M   & 80.56   \\ %
    ResNet 50 + GAM~\cite{Liu2021GlobalAM}   & 149.47M & 81.33  \\ %
    ResNet 50 + GAM (gc*)~\cite{Liu2021GlobalAM}  & 57.05M & 81.01  \\ %
    ResNet 50 + EncodeNet  & 23.71 M  & 80.10  \\ %
    \bottomrule
  \end{tabular}
\end{table}
Attention mechanism based techniques increase the performance of the DNN models, but that usually leads to a substantial increase in the number of model parameters.
We compared EncodeNet with different attention mechanism techniques, namely Squeeze-and-Excitation Networks (SE) ~\cite{hu2018squeeze}, Bottleneck Attention Module (BAM)~\cite{park2018bam}, Convolutional block attention module (CBAM) \cite{woo2018cbam}, and Global attention mechanism (GAM) \cite{Liu2021GlobalAM}.
In Table \ref{table:attention_mechanism}, we observe that our framework can increase the accuracy of baseline DNN at par with other attention mechanism, but without increasing the model parameter size.
For example, Global attention Mechanism (GAM) ~\cite{Liu2021GlobalAM} shows higher improvement in classification accuracy (81.33\%) compared to EncodeNet (80.10\%), but significantly increases the model parameter size (approx 5 times).

\section{Conclusion}
\label{Conclusion}


This paper introduces a novel framework to improve the classification accuracy of baseline deep neural network (DNN) models without introducing additional parameters. Our proposed methodology outperforms recent knowledge distillation and attention mechanism approaches focused on enhancing DNN model efficacy. For each given baseline DNN model, we utilize the feature extraction layer as the encoder module of an autoencoder, design the decoder based on this feature extractor, and subsequently retrain the autoencoder to convert the given images into representative images (easy images). For robust training of autoencoder, we also implemented intraclass clustering that helped to minimize the reconstruction loss. Furthermore, we integrate the concept of knowledge transfer by employing the frozen encoder section, appending the baseline classifier, and retraining it for classification tasks. Our methodology is validated using CIFAR-10 and CIFAR-100 datasets, showcasing its capability to boost the performance of any DNN model. Additionally, we conducted a comparative analysis with existing techniques that concentrate on enhancing baseline DNN performance. 


{
    \bibliographystyle{splncs04}
    \bibliography{main}
}

\end{document}